\documentclass{article}

\usepackage[arxiv,main]{neurips_2026}

\usepackage[utf8]{inputenc} 
\usepackage[T1]{fontenc}    
\usepackage{hyperref}       
\usepackage{url}            
\usepackage{booktabs}       
\usepackage{amsfonts}       
\usepackage{nicefrac}       
\usepackage{microtype}      
\usepackage{xcolor}         
\usepackage{graphicx}
\usepackage{amsmath}
\usepackage{multirow}
\usepackage{wrapfig}

\definecolor{Orange}{RGB}{255,165,0}
\newcommand{\ziyang}[1]{\textcolor{black}{#1}}

\newcommand{\ourwork}{\textbf{AtlasVid~}}
\title{\ourwork: Efficient Ultra-High-Resolution Long Video Generation via Decoupled Global-Local Modeling}

\author{%
  Ziyang Mai\thanks{Co-first authors, equal contribution.}\\
  Dartmouth College \\
  \texttt{ziyang.mai.gr@dartmouth.edu} \\
  \And
  Yuyao Zhang\footnotemark[1]\\
  Dartmouth College \\
  \texttt{yuyao.zhang.gr@dartmouth.edu} \\
  \And
  Yu-Wing Tai \\
  Dartmouth College \\
}

\begin{document}

\vspace{-4in}
\maketitle
\begin{figure}[h]
    \centering
    \includegraphics[width=\linewidth]{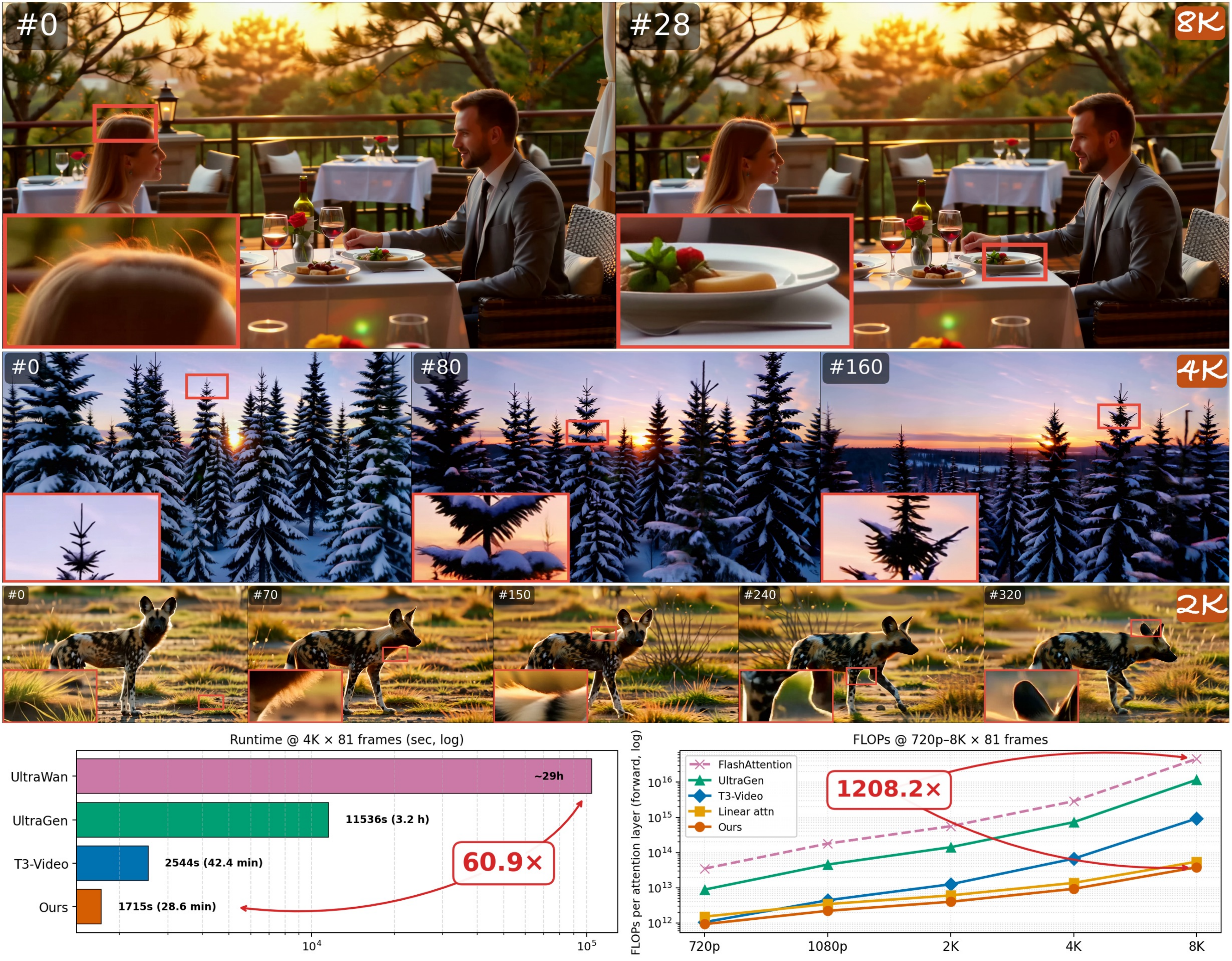}
    \caption{\ourwork enables the generation of ultra-high-resolution and long-duration videos in different settings, including 8K 29 frames, 4K 161 frames and 2K 321 frames. Frame index indicated in the top-left corner and the output resolution in the top-right corner. \textbf{Bottom:} AtlasVid runs 60.9× faster than UltraWan at 4K × 81 frames (left) and reduces per-layer attention FLOPs by up to 1208.2× over FlashAttention from 720p to 8K (right).}
    \label{fig:Teasor}
\end{figure}

\begin{abstract}
Recent diffusion-based video generators have achieved remarkable visual fidelity and prompt controllability, yet scaling them to ultra-high-resolution (UHR) long videos remains prohibitively expensive. The difficulty is especially pronounced for long single-shot generation where a continuous scene must preserve global temporal coherence, and fine-grained spatial details without relying on clip transitions or autoregressive shot stitching. In this work, we revisit this challenge from the perspective of decoupled modeling. We argue that existing video diffusion models already encode strong local visual priors, while the main bottleneck lies in efficiently extending global spatiotemporal modeling as resolution and duration increase. Based on this insight, we propose~\ourwork, a decoupled global-local framework for efficient UHR long video generation. ~\ourwork first generates a low-resolution and low-FPS global semantic proxy via temporally scaled RoPE, thereby extending the temporal horizon without increasing the training token count. 
Guided by this proxy, a high-resolution detail branch performs joint denoising with hierarchical locality-preserving attention.
Reordered spatiotemporal windows preserve geometric locality and asymmetric global-local attention injects aligned semantic guidance and preserves the model's pretrained ability. This design enables resolution-agnostic training: the model is trained only at 720P with lightweight LoRA adaptation, yet generalizes directly to 4K and beyond for longer (>10s) video synthesis. 
Experiments show that~\ourwork substantially improves the efficiency of ultra-high-resolution long video generation, achieving high-quality UHR long video generation with $60.9\times$ speed up and significantly less training cost and even better performance than native 4K video generators.
\end{abstract}

\section{Introduction}
\label{sec:intro}

The field of video generation has advanced rapidly, driven by Diffusion Models~\citep{ho2020denoising} and the Diffusion Transformer (DiT) architecture~\citep{peebles2023scalable}. Recent text-to-video (T2V) systems~\citep{arkhipkin2025kandinsky, chen2025skyreels, li2026skyreels, kong2024hunyuanvideo, hunyuanvideo2025, lin2024open, HaCohen2024LTXVideo, hacohen2026ltx, ma2025step, wan2025wan} have significantly improved realism and visual fidelity, enabling compelling video synthesis from text or image prompts. As these models mature, the focus is shifting from prompt following and visual quality toward high-resolution, long-duration generation, motivated by applications in filmmaking, television, and professional content creation, where fine-grained spatial detail and long-range temporal coherence are both critical. To meet this demand, industrial systems~\citep{openai2025sora2, google2025veo31, bytedance2026seedance2, kuaishou2026kling3} rely on massive data and computation to support generation up to 2K resolution, typically in a clip-by-clip autoregressive manner. However, each clip is usually limited to less than 5 seconds. 

Meanwhile, academic works have explored long-video streaming generation~\citep{yin2024slow, huang2025self, chen2026context, teng2025magi} and planning-based methods for extended video creation~\citep{huang2024context, zheng2024videogen, zhao2024moviedreamer, guo2025long}, yet genuinely long single-shot generation remains out of reach. 

In real-world scenarios, however, high-resolution long single-shot videos are often required, such as cinematic long takes, music and dance performances, and documentary-style footage.

A major challenge is the inherent complexity of the DiT architecture, whose full-attention mechanism scales quadratically with the spatiotemporal input size, i.e., $O((HWT)^2)$, where $H$, $W$, and $T$ denote the height, width, and temporal length of the video. Consequently, doubling all three dimensions increases the computational cost by $64\times$. 

Moreover, directly training models at ultra-high resolution can introduce additional and unpredictable difficulties such as severe memory pressure, training instability at extremely long token sequences, and the scarcity of native 4K long-video training data.
Existing methods~\citep{he2024venhancer, xie2025star,zhuang2025flashvsr} alleviate this issue with a "generation-then-super-resolution" pipeline. However, this "pseudo high-resolution" paradigm mainly improves sharpness and often fails to recover sufficiently rich high-frequency details. Another line of work~\citep{chen2025sana, zhang2025spargeattn, wang2025lingen} improves efficiency through sparse or linearized attention mechanisms, although faster, they do not fundamentally resolve the substantial data bottleneck of high-resolution long-video generation.

These limitations prompt a rethinking of what fundamentally restricts high-resolution long-video generation. We hypothesize that pretrained video generation models already possess much of the visual knowledge required to synthesize plausible fine-grained details at higher resolutions, since pretraining exposes them to similar objects, structures, and scenes across different visual scales. From this perspective, the main bottleneck is not an inherent lack of high-resolution generation capacity, but the difficulty of modeling long-range dependencies efficiently as the spatiotemporal resolution grows.

Motivated by this view, we propose a hierarchical locality-preserving attention mechanism for efficient high-resolution long-video generation. Our design decouples the modeling of local neighborhood structure from global semantics, allowing the model to scale more effectively across resolutions. Furthermore, we introduce a temporal-scale RoPE strategy for long-video modeling, which enlarges the temporal modeling range without increasing the token count. By combining hierarchical locality-preserving attention with temporal-scale positional modeling, our framework enables efficient ultra-high-resolution video generation and reduces the data and resource demands of scaling.

\begin{itemize}
    \item Through the decoupled modeling of global semantic and local details, we significantly reduce the computing complexity of ultra-high-resolution generation leading to a speedup of $60.9\times$ compare to Wan2.1-1.3B baseline.
    \item Enabled by our resolution-agnostic training paradigm, ~\ourwork is the first method to jointly scale up both spatial resolution and temporal duration (i.e 4K, 321frames) without the data bottleneck to our knowledge.
\item The full pipeline can be trained via LoRA fine-tuning at 720P resolution on as few as \textbf{2 NVIDIA RTX Pro 6000 GPUs} and inference on 1 GPU within 29 minutes, with the learned capability transferring directly to 4K inference without any additional high-resolution training stage. In comparison to other method which requires massive training on more than 32 GPUs clusters,~\ourwork substantially lowers the resource barrier ultra-high-resolution long video generation.
\end{itemize}

\section{Related Work}
\begin{table}[t]
\centering
\caption{Training resource comparison with state-of-the-art high-resolution video generation methods.
}
\label{tab:training_cost}
\setlength{\tabcolsep}{6pt}
\resizebox{\columnwidth}{!}{%
\begin{tabular}{lcccccc}
\toprule
\textbf{Method} & \textbf{GPUs} & \textbf{Training Stage} & \textbf{Max Frames} & \textbf{Max Reso.} & \textbf{Duration} \\
\midrule
UltraWan-4K~\citep{xue2025ultravideo}  & 7.6K H20-hours & LoRA on Wan-1.3B   & 29f  & 4K  & $\sim$1.2s \\
UltraGen~\citep{hu2026ultragen}        & 32$\times$H20  & 4K direct fine-tuning              & 29f  & 4K  & $\sim$1.2s \\
T3-Video~\citep{zhang2025transform}    & 64             & 720P $\rightarrow$ 4K (two-stage)  & 81f  & 4K  & $\sim$3.4s \\
\midrule
\textbf{\ourwork (Ours)} & \textbf{2$\times$RTX Pro 6000} & \textbf{720P (+ optional 4K)} & \textbf{321f} & \textbf{$>$4K} & \textbf{$\sim$13.4s} \\
\bottomrule
\end{tabular}
}
\end{table}
\subsection{High Resolution Video Generation}
Recent diffusion transformer based video generation models~\citep{wan2025wan,hunyuanvideo2025,hong2022cogvideo} have demonstrated impressive synthesis quality, yet they are still largely trained at resolutions up to 720P. Scaling these models to ultra-high resolution (UHD), such as 4K, remains challenging due to the quadratic complexity of full attention with respect to token count and the scarcity of high-quality 4K video data. 
Training-free methods~\citep{he2023scalecrafter,zhang2024hidiffusion,qiu2025cinescale,wu2025freeswim} adapt pretrained models to higher resolutions by modifying attention patterns, receptive fields, or positional encodings at inference time. 
While computationally convenient, they rely entirely on low-resolution priors and often suffer from semantic repetition, over-smoothed textures, and content inconsistency at 4K. 
Video super-resolution methods~\citep{he2024venhancer,xie2025star,zhuang2025flashvsr} instead cascade low-resolution generation with a dedicated spatial upscaler, but such pipelines are mainly restricted to low-level texture enhancement and cannot reliably correct semantic errors or synthesize genuinely new high-frequency content. 
Native high-resolution fine-tuning approaches~\citep{xue2025ultravideo,hu2026ultragen,zhang2025transform,zhao2026luve} directly adapt foundation models on curated high-resolution datasets. However, these methods primarily scale spatial resolution and remain limited to short clips. 
In contrast, our proposed~\ourwork is resolution-agnostic: it scales effectively to higher resolutions without relying heavily on native 4K training data.
\paragraph{Discussion with similar works.}
\textbf{\textit{i)} Methodology.} UltraWan, UltraGen, and T3-Video all rely on large-scale 4K data for training. UltraGen adopts carefully designed hierarchical attention modules to reduce computation, while T3-Video uses transformed window attention for more efficient training and inference. In contrast, our method decouples global-local modeling and introduces locality-preserving attention, enabling efficient 4K inference while requiring training only at 720P with optional 4K finetuning for more realistic results. 
\textbf{\textit{ii)} Training resources.} As shown in Table~\ref{tab:training_cost}, UltraGen uses 32 H20 GPUs, and T3-Video requires 64 GPUs for training, whereas our method requires only 2 GPUs. 
\textbf{\textit{iii)} Results.} Our method is the first to enable long single-shot 4K video generation.


\subsection{Efficient Video Generation} Efficient video generation has attracted increasing attention due to the heavy computational cost of diffusion-based video models, which becomes more prohibitive at ultra high spatial temporal resolutions. Existing methods mainly improve efficiency through several complementary directions. First, hardware-aware attention kernels such as FlashAttention~\citep{dao2023flashattention,shah2024flashattention} and SageAttention~\citep{zhang2024sageattention,zhang2024sageattention2,zhang2025sageattention3} accelerate standard attention computation and reduce memory overhead without changing the model architecture.
Second, caching-based methods exploit redundancy across denoising timesteps by reusing intermediate features or attention outputs, thereby reducing repeated computation during sampling~\citep{liu2024timestep,ma2025magcache,fan2025taocache}.
Third, many recent approaches improve efficiency at the architectural level by introducing complexity-reduced spatiotemporal modeling, such as training-free sparsification, trainable sparse attention, and linear-complexity designs~\citep{zhang2025spargeattn,wang2025lingen, zhang2025vsa}.
In addition, distillation-based methods reduce the number of sampling steps for faster deployment~\citep{luo2023latent,yin2024one}.
Different from these works that primarily target general video generation efficiency, our work focuses on native high-resolution video generation and develops sparse-attention-based optimization tailored to low reference guidance setting.

\begin{figure}[t]
    \centering
    \includegraphics[width=\linewidth,trim={0 0 0.1cm 0}, clip,trim={0 0 0.1cm 0}, clip]{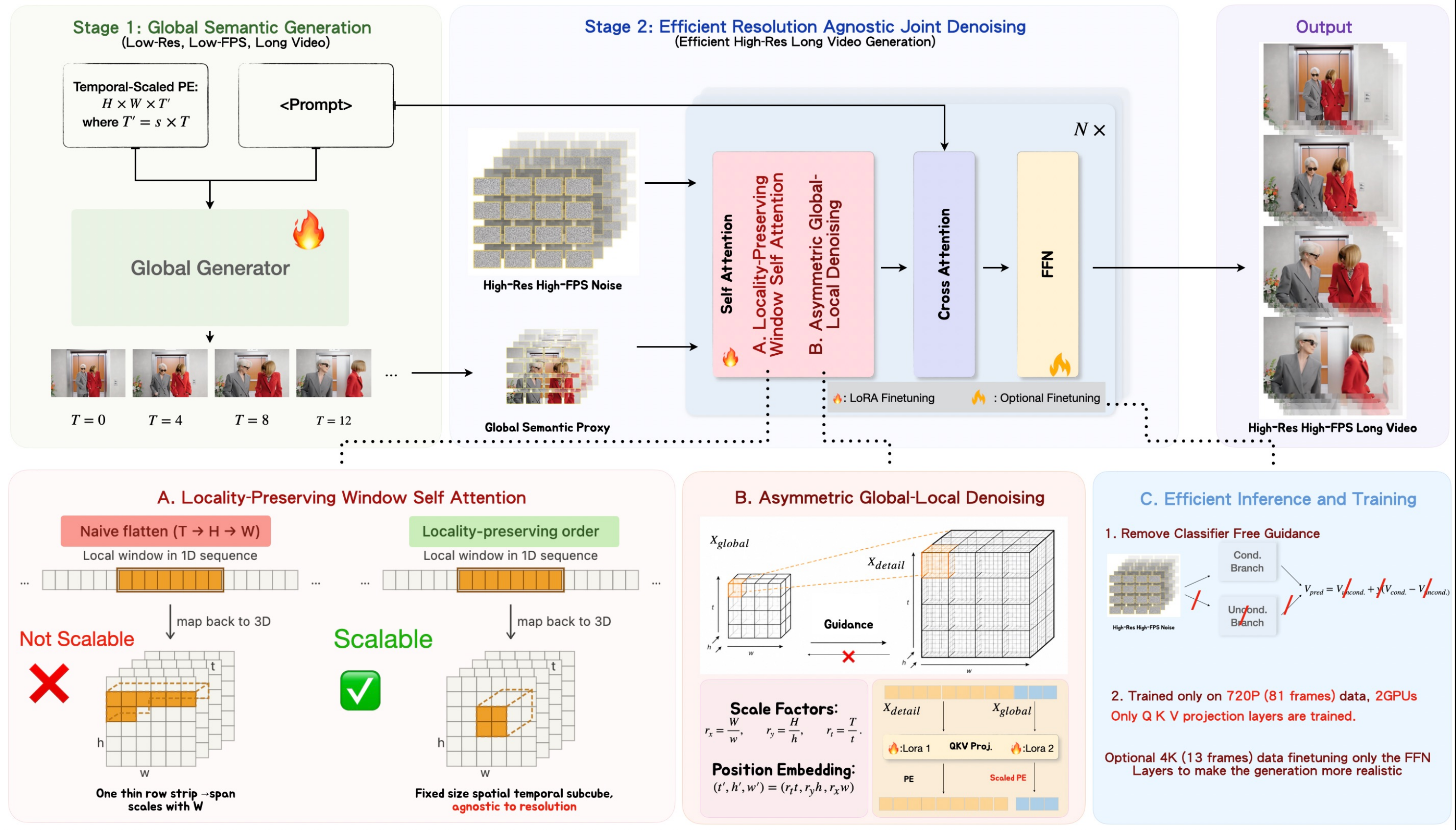}
    \caption{Pipeline of~\ourwork. It first employs a semantic generator to produce a low-resolution, low-frame-rate video that serves as a global semantic proxy. Conditioned on this reference, the second stage performs spatiotemporal detail generation through an efficient hierarchical locality-preserving attention mechanism, enabling ultra-high-resolution long-video synthesis(UHRL video) with substantially improved computational efficiency. Section A, B, and C demonstrate the detailed designs for scalable UHRL video generation.}
    \vspace{-0.2in}
    \label{fig:pipeline}
\end{figure}
\section{Method}
To generate high-resolution long single-shot videos without incurring the prohibitive quadratic cost of standard self-attention or the substantial training burden of direct high-resolution long-video modeling, \ourwork~reformulates spatiotemporal video generation as an efficient hierarchical two-stage pipeline. As illustrated in Figure~\ref{fig:pipeline}, the first stage employs a semantic generator to produce a low-resolution, low-frame-rate video that serves as a global semantic proxy. Conditioned on this reference, the second stage performs spatiotemporal detail generation through an efficient hierarchical locality-preserving attention mechanism, enabling high-resolution long-video synthesis with substantially improved computational efficiency. In the remainder of this section, Section~\ref{sec-preliminaries} introduces the preliminaries of DiT-based video diffusion models, Section~\ref{sec-decoupled_modeling} presents our decoupled formulation of global semantic and local detail modeling for scalable and efficient generation, and Section~\ref{sec-joint-denoising} describes how the global semantic proxy guides high-resolution long-video denoising.

\subsection{Preliminaries of DiT based Video Diffusion Models}
\label{sec-preliminaries}
Most recent DiT-based video generation models adopt a full-attention Transformer architecture to model spatiotemporal dependencies in latent video representations. Given an input video, a 3D-VAE first encodes it into a latent tensor of shape $D \times T \times H \times W$, where $D$ is the channel dimension and $T,H,W$ denote the temporal and spatial dimensions. The latent tensor is then patchified and flattened into a 1D token sequence of length $N = T H W$. Full self-attention is applied over the entire sequence, producing an $N \times N$ attention map with complexity $O(N^2 D)$. As a result, the cost grows quadratically with the spatiotemporal token count, which quickly becomes prohibitive for high-resolution and long video generation.

\subsection{Decoupled Modeling of Global Semantics and Local Details}
\label{sec-decoupled_modeling}
Generating high-resolution long videos requires simultaneously modeling long-range spatiotemporal dependencies over extremely long token sequences while preserving fine-grained local fidelity. We observe that recent state-of-the-art video generation models already demonstrate strong visual synthesis ability across diverse objects, scales, and motion patterns, suggesting that pretrained models have largely internalized the local visual priors required for detail generation. The main challenge, instead, lies in efficiently maintaining coherent global semantics as the spatiotemporal extent grows. Based on this observation, we decouple global semantic modeling from local detail generation and address them separately in our framework.

\noindent\textbf{Global Semantic Generation via Temporally Scaled Positions.} 
The pretrained base model already possesses the capacity to generate low-resolution semantic proxies, but extending such semantic modeling to longer temporal horizons still requires adaptation. A straightforward solution is to finetune the base model for longer video generation directly. However, this quickly becomes prohibitively expensive due to the quadratic complexity of self-attention with respect to temporal token length. 

Importantly, global semantic modeling does not require the full frame rate of the target video. Instead, a low-frame-rate video is sufficient to capture the coarse scene evolution and long-range temporal semantics. Based on this observation, we construct the first stage as a low-frame-rate semantic generator. Specifically, we enlarge the temporal indices used in RoPE by a factor of $r_t$, such that adjacent generated frames are interpreted as being separated by larger temporal intervals. This allows the model to represent a longer temporal horizon using the same number of frames. For example, when $r_t = 4$, a 4fps 20-second semantic proxy has the same temporal token length as an original 5-second video, reducing the self-attention cost by $16\times$ compared with directly modeling the same duration at the original frame rate. We implement this adaptation by finetuning the base model with LoRA, enabling efficient learning of long-horizon semantic generation while preserving the pretrained visual prior.

\noindent\textbf{Locality-Preserving Efficient Attention for Detail Generation.}
Let the target video contain $N$ latent tokens. Applying full self-attention over all tokens leads to a quadratic complexity of $O(N^2)$, which is prohibitive for high-resolution long-video generation. Existing efficient attention schemes typically reduce this cost by restricting attention to local windows, yielding a complexity of $O(Nb)$ with window size $b$. However, when windows are formed directly on the flattened token sequence, sequence locality does not necessarily coincide with geometric locality in the original spatiotemporal video volume. As a result, tokens grouped within the same attention window may not belong to the same coherent local region in space and time.

We therefore introduce a locality-preserving efficient attention mechanism for detail generation. Figure~\ref{fig:pipeline}~(A) demonstrates the differences between ours and naive local-attention. Specifically, given a latent video volume of size $T \times H \times W$, we partition it into spatiotemporal cubes of size $t \times h \times w$, chosen such that each cube remains within the local generation regime that the pretrained base model can effectively handle. We then reorder the flattened token sequence so that tokens from the same cube become contiguous. Window attention is applied on the reordered sequence with window size $b = thw$, ensuring that each sparse attention block corresponds to an actual local spatiotemporal neighborhood. 

This construction preserves the computational efficiency of local attention while making the sparse attention pattern compatible with the geometric structure of the video. Consequently, it better preserves the pretrained model's ability to synthesize fine local details. In practice, we additionally allow attention across adjacent cubes to reduce blocking artifacts near cube boundaries.

\subsection{Efficient Resolution Agnostic Joint Denoising}
\label{sec-joint-denoising}
Given the global semantic proxy $X_{\text{global}}$ with length $n$ and the target noisy video tokens $X_{\text{detail}}$ with length $N$, our next goal is to construct an efficient joint denoising architecture that allows $X_{\text{global}}$ to guide the generation of $X_{\text{detail}}$ while remaining scalable across resolutions.

\noindent\textbf{Efficient Joint Denoising with Hierarchical Attention.}
To inject global semantic guidance into high-resolution detail generation, we concatenate the two token streams as
$X = [X_{\text{detail}}; X_{\text{global}}]$,
and perform joint self-attention followed by text cross-attention. The key design principle is an asymmetric hierarchical attention pattern in which the coarse global proxy provides semantic guidance, while the high-resolution detail branch performs denoising under both local detail interactions and global semantic conditioning.

In self-attention, $X_{\text{global}}$ attends only to itself, while $X_{\text{detail}}$ attends both to its own tokens through the locality-preserving attention described in Section~\ref{sec-decoupled_modeling} and to the aligned global semantic tokens. The self-attention is formulated as
\[
\mathrm{SelfAttn}(Q,K,V)
=
\mathrm{Softmax}\!\left(\frac{QK^{\top} + M}{\sqrt{d}}\right)V, M=
\begin{bmatrix}
M_{dd} & M_{dg}\\
-\infty & M_{gg}
\end{bmatrix} \in \mathbb{R}^{(N+n)\times(N+n)}
\]
Here, $M_{dd}$ denotes the locality-preserving mask for detail-to-detail attention, $M_{dg}$ denotes the mask for detail-to-global attention, and $M_{gg}$ denotes unrestricted self-attention among global tokens. The $-\infty$ block prevents global tokens from attending to detail tokens, thereby enforcing one-way semantic guidance from the global proxy to the noisy detail branch. For text cross-attention, we only allow $X_{\text{detail}}$ to attend to text tokens. This asymmetric design ensures that the global semantic proxy remains clean and is not corrupted by noisy detail features during denoising. 

To adapt the pretrained DiT backbone to this new joint denoising pattern, we apply LoRA finetuning to all the DiT layers associated with global semantic conditioning and to only the query, key, and value projection layers in the detail branch. In contrast, we keep the feed-forward layers in the detail branch frozen in this joint denoising pattern learning process, so that the pretrained model's original local generative capability is preserved as much as possible while learning the detail-to-global mapping. 
As an optional refinement stage, we additionally finetune previously frozen FFN layers using a small and short (13-frame) videos to further improve local realism.

\noindent\textbf{Scaled Spatial-Temporal RoPE for Global-Local Matching.}
To make the global semantic proxy provide accurate guidance for high-resolution detail generation, we align its positional encoding with the coordinate system of the target video. Demonstrated in Figure~\ref{fig:pipeline}~(B), for a global proxy token located at spatial-temporal index $(w',h',t')$, we scale its RoPE coordinates by
\[
r_x=\frac{W}{w}, \qquad r_y=\frac{H}{h}, \qquad r_t=\frac{T}{t},
\]
so that its position is mapped to the corresponding location in the high-resolution long-video latent grid. In this way, each token in $X_{\text{global}}$ serves as an anchor for the corresponding spatiotemporal cube in $X_{\text{detail}}$, enabling consistent global-to-local semantic guidance during denoising.

This positional alignment naturally complements the asymmetric hierarchical attention described above: the global proxy provides semantically aligned coarse guidance, while the detail branch focuses on synthesizing local high-frequency content within each spatiotemporal neighborhood. Since our framework explicitly decouples global semantic modeling from local detail generation, training only needs to learn the new locality-preserving attention pattern and the hierarchical global-local correspondence, rather than directly modeling the full target spatiotemporal scale end-to-end. 

As a result, \ourwork~does not require long ultra-high-resolution videos (e.g., 4K, 321-frame videos) for training as illustrated in Figure~\ref{fig:pipeline}(c) , making the overall training pipeline resolution-agnostic and scalable. Moreover, due to the existence of the global proxy as condition, we remove the Classifier-Free Guidance which significantly speeds up the inference.

\begin{figure*}[t]
    \centering
    \includegraphics[width=\linewidth]{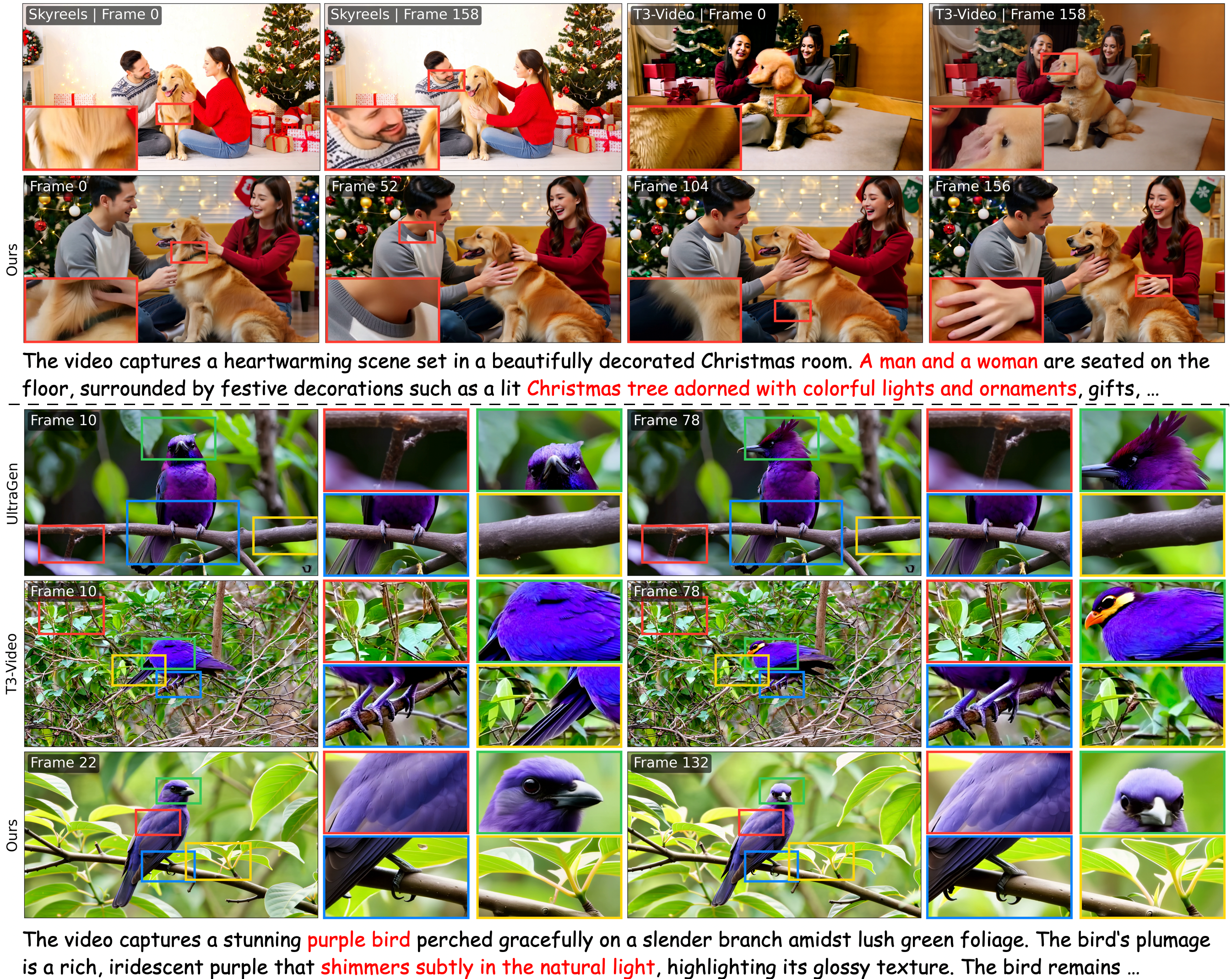}
    \vspace{-0.15in}
    \caption{
    Qualitative comparison of long ultra-high-resolution video generation. Top: The first two rows compare \ourwork with SkyReels-V2 at 720P and T3-Video at 4K. All three methods generate 161-frame results. T3-Video exhibits structural artifacts, color shifts, and quality degradation when extended to 161 frames, whereas \ourwork maintains coherent structures and stable visual quality. Bottom: Rows 3–5 show the 81-frame comparison with UltraGen at 1080P and T3-Video at 4K. \ourwork preserves 4K-level details and is the only method that scales reliably beyond 81 frames.
    }
    \vspace{-0.15in}
    \label{fig:qualitative_uhrl}
\end{figure*}

\section{Experiments}
\subsection{Implementation Details}
\noindent\textbf{Training details.} We implement~\ourwork on top of Wan2.1-1.3B for both stage 1 and stage 2. Models are trained on UltraVideo using AdamW optimizer for 15K iterations (Batch size 1) on 2 RTX 6000 Ada GPUs with gradient accumulation 4. Learning rates is 1e-4 with LoRA rank 16. Models are trained on 720p, 81 frames, and is successfully scaled beyond 4K, 321frames. Stage~1 fine-tunes the base model with temporal-scale RoPE ($r_t{=}4$) for long-horizon low frame-rate semantic generation. We set the spatial-temporal block size in stage 2 to be $(256,256,32)$ which is equivalent to $(8,8,4)$ in the latent space and adapts a smaller block-size for the border cases when the spatial-temporal resolution cannot be divided by the block-size. We adapt the mask rule for flex attention to support dynamic length and resolution.

\ziyang{\noindent\textbf{Baselines.}
We compare against: 
Wan2.1-T2V-1.3B~\citep{wan2025wan}, 
UltraWan~\citep{xue2025ultravideo}, UltraGen~\citep{hu2026ultragen}, and
T3-Video-T2V-1.3B~\citep{zhang2025transform}. Note that neither Wan2.1-T2V-1.3B, nor UltraGen have released official 4K-resolution checkpoints, we therefore evaluate them at their highest support resolution, 720p and 1080p separately.} 

\subsection{Qualitative Results}
\noindent\textbf{Demonstration of UHR long videos.} Figure~\ref{fig:Teasor} presents examples of ultra-high-resolution long videos generated by our method, reaching up to 8K resolution and 321 frames in temporal length. These results demonstrate the scalability of our framework in both spatial resolution and temporal duration, while maintaining coherent scene structure and fine-grained visual details.

\noindent\textbf{Comparison on 4K long videos}
Figure~\ref{fig:qualitative_uhrl} provides qualitative comparisons for long 4K video generation. The upper section compares long video generation ability on 161 frames with the long-video generator SkyReels-V2 under 720p and the native 4K generator T3-Video under 4K. Our method is the only one that produces plausible long 4K results. When extrapolated to longer temporal lengths, T3-Video exhibits noticeable quality degradation and prompt misalignment, such as color shift and incorrectly changing the prompted man into a woman. SkyReels-V2 can generate long video with consistency, but their result are limited to 720p. The lower section further compares our method with native UHR video generation methods, including UltraGen and T3-Video. UltraGen produces noticeably inconsistent results when using its released 1080P checkpoint, such as the feather on the head in green box. Though our method and T3-Video preserve detailed structures at 4K resolution, T3-Video still exhibits structural artifacts, such as distorted bird shape in green and yellow box, with extra legs in blue box, due to the lack of explicit global semantic control. In contrast, our decoupled design provides stable semantic guidance and more coherent high-resolution long-video synthesis.

\subsection{Quantitative Results}

Table~\ref{tab:quantitative_4k} reports quantitative results using both VBench and high-resolution evaluation metrics, with detailed metric definitions provided in the appendix. We randomly sample 100 prompts from VBench as our test set. Wan2.1 and UltraGen are evaluated at 720P and 1080P respectively, as their 4K checkpoints are not publicly available, while UltraWan, T3-Video, and \ourwork are evaluated at 4K resolution. \ourwork achieves the best performance on most VBench metrics among existing 4K generation methods, and remains competitive with pretrained models evaluated at 720P resolution. These results demonstrate the strong capability of \ourwork in synthesizing high-quality ultra-high-resolution videos while preserving both visual fidelity and temporal consistency.

\begin{table}[h]
\centering
\caption{Quantitative comparison on 4K long video generation. Our method achieves the best performance among 4K generation methods and remains competitive with 720P base models.}
\label{tab:quantitative_4k}
\scriptsize
\setlength{\tabcolsep}{2.0pt}
\renewcommand{\arraystretch}{1.05}

\resizebox{\linewidth}{!}{
\begin{tabular}{lccc|ccccc|ccc}
\toprule
\textbf{Method}
& \multicolumn{3}{c|}{\textbf{4K HD Metrics}}
& \multicolumn{5}{c|}{\textbf{VBench Motion}}
& \multicolumn{3}{c}{\textbf{VBench Semantics}} \\
\cmidrule(lr){2-4}
\cmidrule(lr){5-9}
\cmidrule(lr){10-12}
& \textbf{HD-FVD}$\downarrow$
& \textbf{HD-LPIPS}$\uparrow$
& \textbf{CLIP}$\uparrow$
& \textbf{SC}$\uparrow$
& \textbf{BC}$\uparrow$
& \textbf{TF}$\uparrow$
& \textbf{MS}$\uparrow$
& \textbf{DD}$\uparrow$
& \textbf{AQ}$\uparrow$
& \textbf{IQ}$\uparrow$
& \textbf{Clr.}$\uparrow$ \\
\midrule
Wan2.1   
& 512.32 & 0.6424 & 0.3076  
& 96.34 & 97.29 & 99.49 & 97.44 & \textbf{85.56}
& 62.43 & 66.51 & 89.58 \\
UltraGen
& 432.71 & 0.6358 & 0.2848
& 94.19 & 96.33 & 98.90 & 98.72 & 71.34
& 58.13 & 65.88 & 97.13 \\
\midrule
UltraWan
& 372.39 & 0.6211 & 0.2903
& 95.71 & \textbf{97.94} & 98.86 & 99.06 & 62.22
& 59.52 & 67.39 & \textbf{99.10} \\
T3-Video
& 311.29 & 0.6471 & 0.2972
& 97.02 & 96.23 & 95.12 & 99.17 & 78.92
& 62.96 & 68.17 & 97.42 \\
\textbf{Ours}
& \textbf{284.71} & \textbf{0.6695} & \textbf{0.3019}
& \textbf{97.41} & 96.54 & \textbf{99.57} & \textbf{99.31} & 82.21
& \textbf{63.31} & \textbf{68.52} & 98.31 \\
\bottomrule
\end{tabular}
}

\vspace{0.45em}
\footnotesize
\end{table}


\noindent\textbf{Efficiency Comparisons.}
We also compare our model's efficiency against the other baselines. The bottom part of Figure~\ref{fig:Teasor} demonstrates the comparison among our attention and other attention mechanisms across 720P to 8K resolutions (81 frames), where our attention even beats linear attention in terms of floating point operations (FLOPs) and surpasses dense attention by 1208.2 times. The right part demonstrates the overall inference time comparison on 4K 81 frames, given the efficient designs we achieve 1.48× faster than T3-Video and 60.9× faster than baseline.
\begin{figure*}[t]
    \centering
    \includegraphics[width=\linewidth]{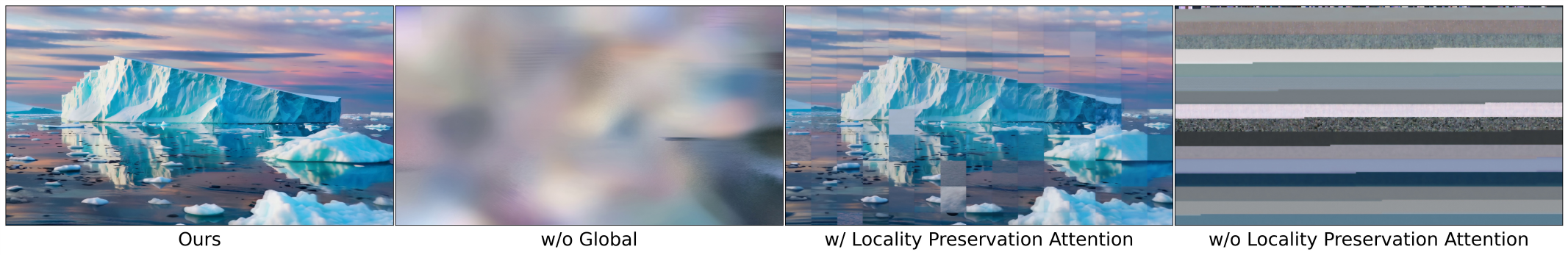}
    \vspace{-0.1in}
    \caption{Ablation on the importance of our attention design. The first two columns demonstrate the effect of global guidance. The third and fourth columns evaluate our locality-preserving attention, where attention to neighboring spatiotemporal blocks is removed to better visualize the learned block structure. All models are trained at 720P and evaluated at 4K.}
    \label{fig:ablation}
\end{figure*}

\section{Ablation Study}
\label{sec:ablation study}

\noindent\textbf{Ablation on different attention patterns.}
We conduct ablation studies on different attention patterns to validate our design choices, with all variants trained only on 720P data. The left part shows that removing the global proxy guidance leads to inconsistent generation results, highlighting the importance of global semantic conditioning. The right part demonstrates the role of locality-preserving attention in extrapolating beyond the training resolution. For clearer visualization, we remove attention to neighboring blocks in this ablation so the "blocks" will be shown in the figure. With our locality-preserving attention, the model successfully scales to 4K generation, whereas naive block attention fails to generalize. These results confirm the effectiveness of our global-local design and locality-preserving attention mechanism.

\begin{wrapfigure}{r}{0.30\linewidth}
    \vspace{-1.2em}
    \centering
    \includegraphics[width=\linewidth]{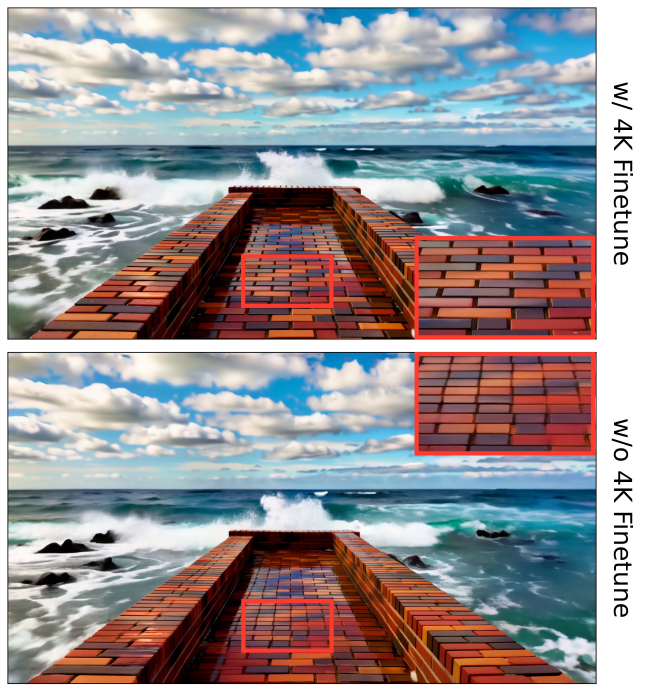}
    \caption{Ablation study on 4K data finetuning. With 4K finetuning (top), 
    the model produces more realistic fine-grained details, 
    while without 4K finetuning (bottom) it can still generate plausible 
    details, demonstrating the robustness of our base model.}
    \label{fig:ablation_4k_finetune}
    \vspace{-1.0em}
\end{wrapfigure}

\noindent\textbf{Ablation on 4K data fine-tuning.}
Figure~\ref{fig:ablation_4k_finetune} compares videos generated by models trained with and without additional 4K real data. Even without 4K fine-tuning, our model already produces clear 4K textures, demonstrating the resolution extrapolation ability of our framework. Additional 4K fine-tuning further improves naturalness and local realism, and is therefore used only as an optional refinement stage. Importantly, this stage is substantially lighter than the native 4K training required by UltraGen and T3-Video: instead of using long 4K 81-frame videos, we only require short 4K 13-frame clips to adapt the feed-forward layers toward more natural high-resolution pixel synthesis.

\paragraph{Limitations.}
Our framework relies on the Stage-1 semantic generator to produce the global proxy, and thus errors or artifacts introduced at this stage may be inherited by the final high-resolution output. In addition, although our decoupled design substantially improves scalability, the final quality still depends on the alignment between the low-resolution proxy and the high-resolution detail branch. In the future we may try larger scale training on more resources and larger dataset to obtain even better results. Our method is also orthogonal to other acceleration techniques like distillation, which may further speed up our method.
\section{Conclusion}
We presented \ourwork, an efficient framework for ultra-high-resolution long video generation. By decoupling global semantic modeling from local detail synthesis, \ourwork avoids directly applying full attention to prohibitively large spatiotemporal token sequences. A low-resolution, low-frame-rate semantic proxy captures long-range scene evolution, while a high-resolution detail branch performs coordinate-aligned joint denoising with hierarchical locality-preserving attention. This design enables resolution-agnostic training: with lightweight LoRA adaptation mainly at 720P, \ourwork generalizes to 4K and beyond while substantially reducing computation. Experiments demonstrate that \ourwork produces coherent, detailed ultra-high-resolution long videos with significantly improved efficiency, offering a scalable path toward accessible ultra-high-resolution video generation.

{
    \bibliographystyle{plainnat}
    \bibliography{main}
}

\clearpage
\appendix

\section{More Results}
\label{sec:More_Results}
\begin{figure}
    \centering
    \includegraphics[width=1\linewidth]{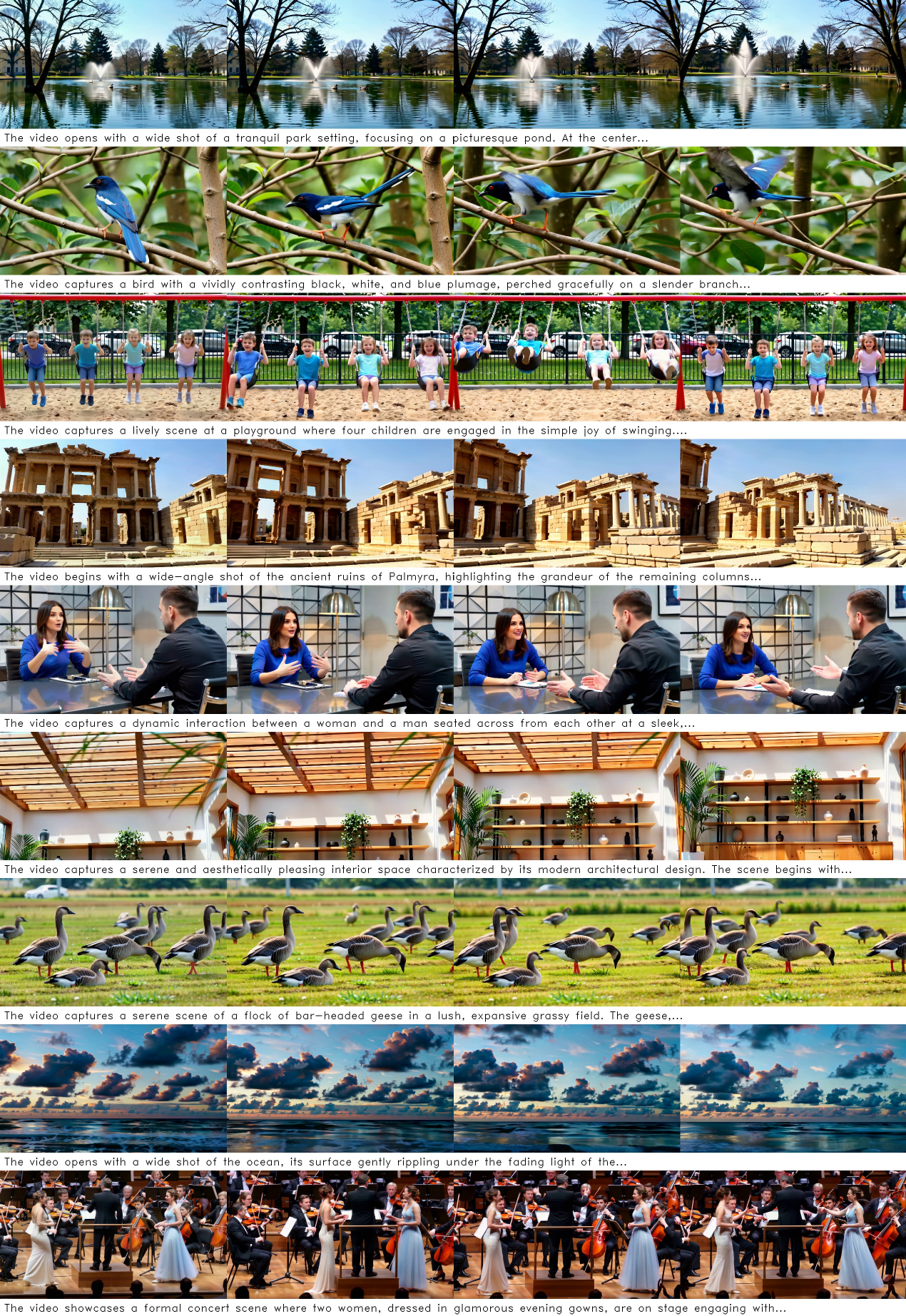}
    \caption{4K 161 Frames results: Each results spans one row. Examples show no quality degradation or color drift. The frame indices are 0, 53, 106, and 160.}
    \label{fig:4k-more}
\end{figure}

\begin{figure}
    \centering
    \includegraphics[width=1\linewidth]{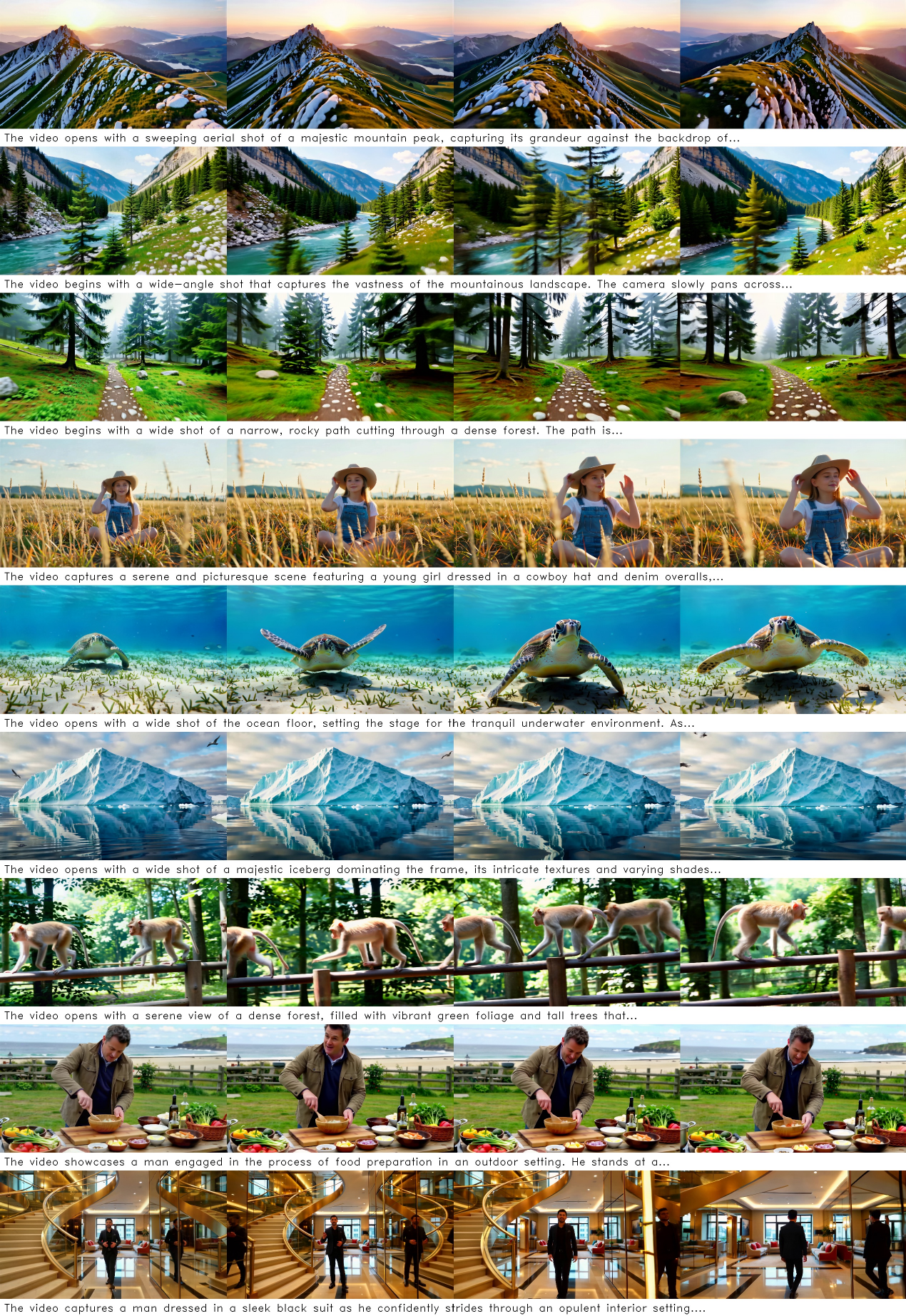}
    \caption{4K 161 Frames results: Each results spans one row. Examples show no quality degradation or color drift. The frame indices are 0, 53, 106, and 160.}
    \label{fig:4k-more2}
\end{figure}

Here we presents more results on 4K 161 frames, 2K 321 frames and 8K 29 frames in Figure~\ref{fig:4k-more}, Figure~\ref{fig:4k-more2}, Figure~\ref{fig:2k-more}, Figure~\ref{fig:2k-more2} and Figure~\ref{fig:8k-more}. We will include the videos in the supplementary materials.

\begin{figure}
    \centering
    \includegraphics[width=1\linewidth]{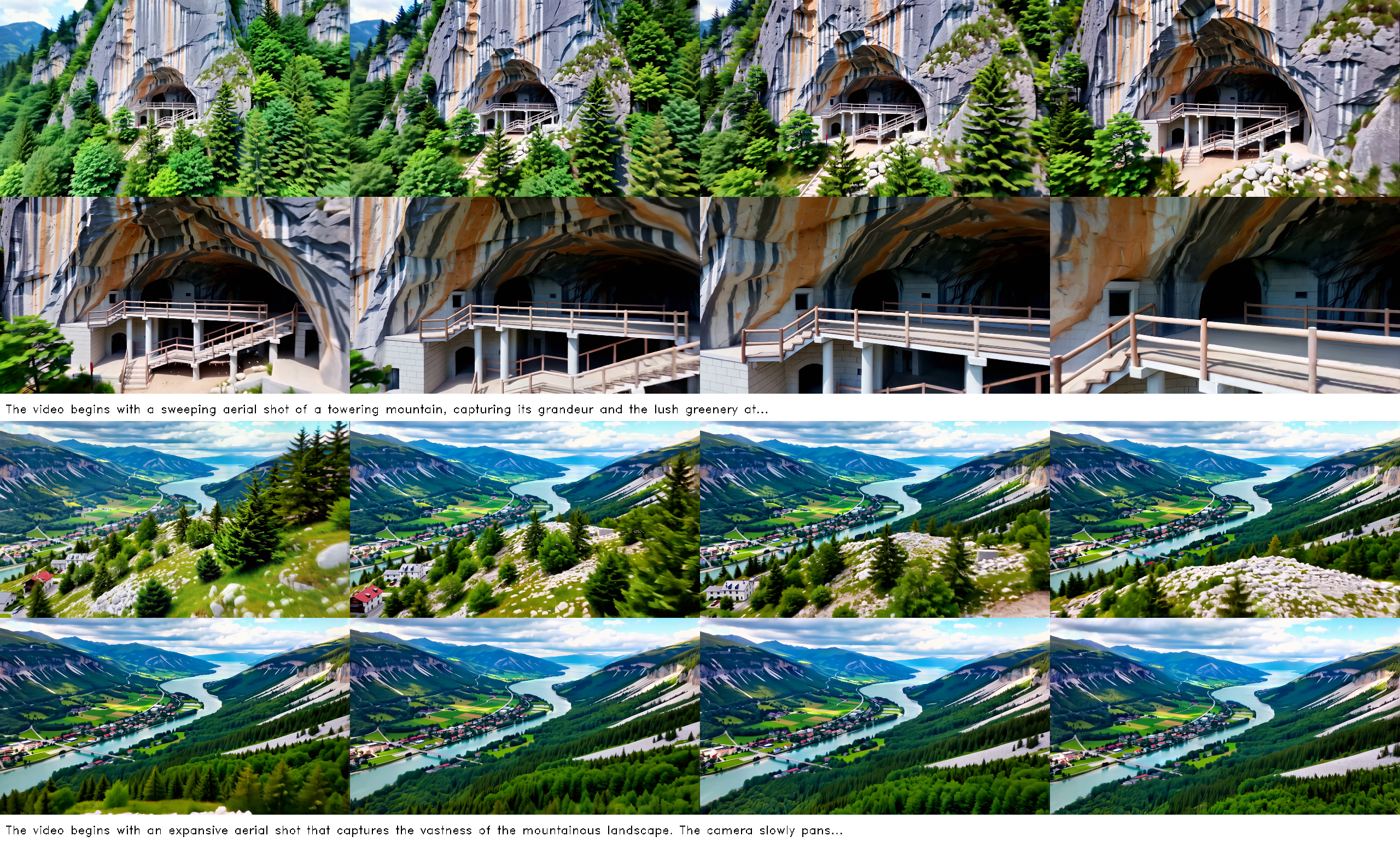}
    \caption{2K 321 Frames results: Each results spans for two rows. The examples here shows large continuous camera movements. The frame indices are 0, 45, 91, 137, 182, 228, 274 and 320.}
    \label{fig:2k-more}
\end{figure}

\begin{figure}
    \centering
    \includegraphics[width=1\linewidth]{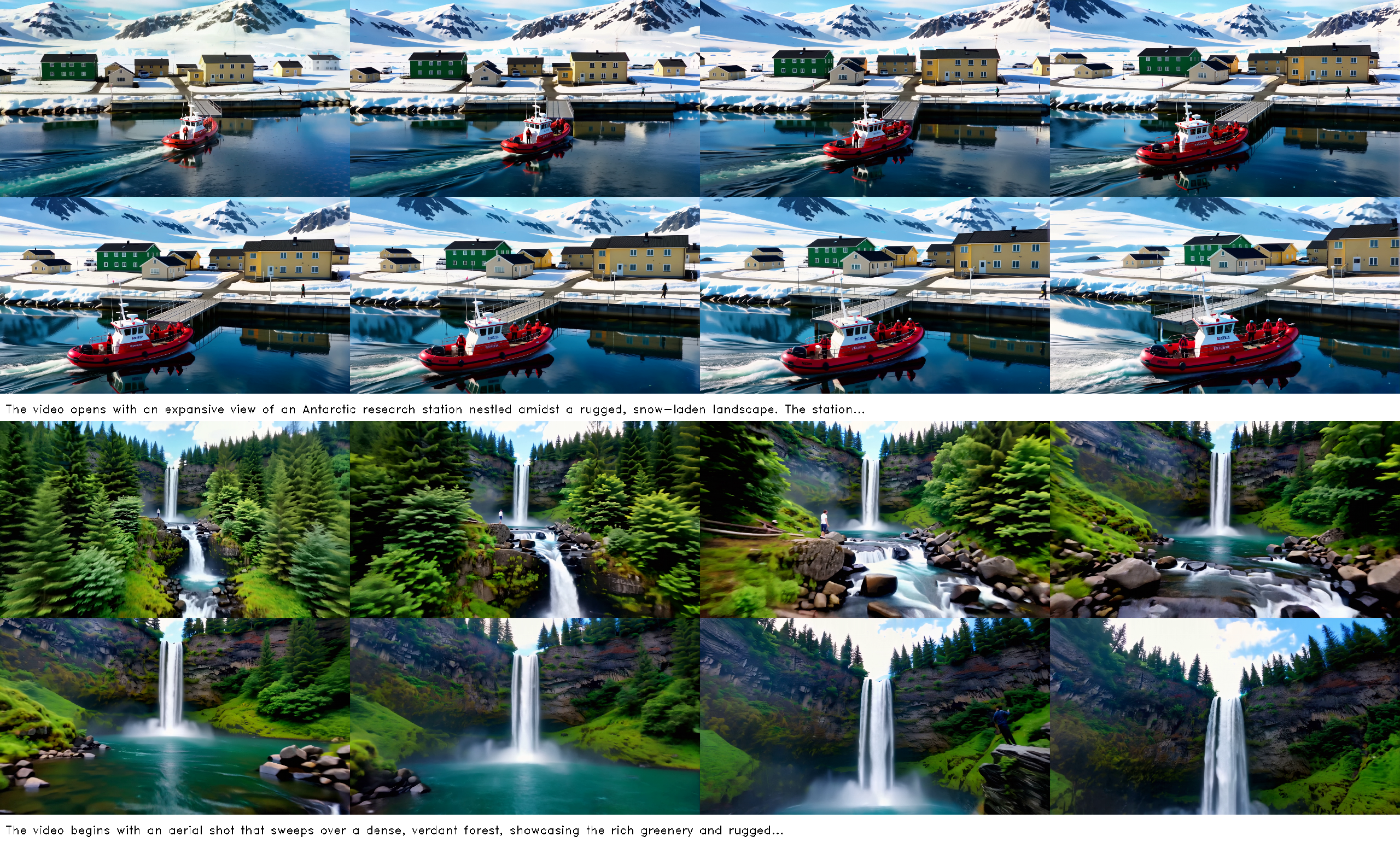}
    \caption{2K 321 Frames results: Each results spans for two rows. The examples here shows large continuous camera movements. The frame indices are 0, 45, 91, 137, 182, 228, 274 and 320.}
    \label{fig:2k-more2}
\end{figure}

\begin{figure}
    \centering
    \includegraphics[width=1\linewidth]{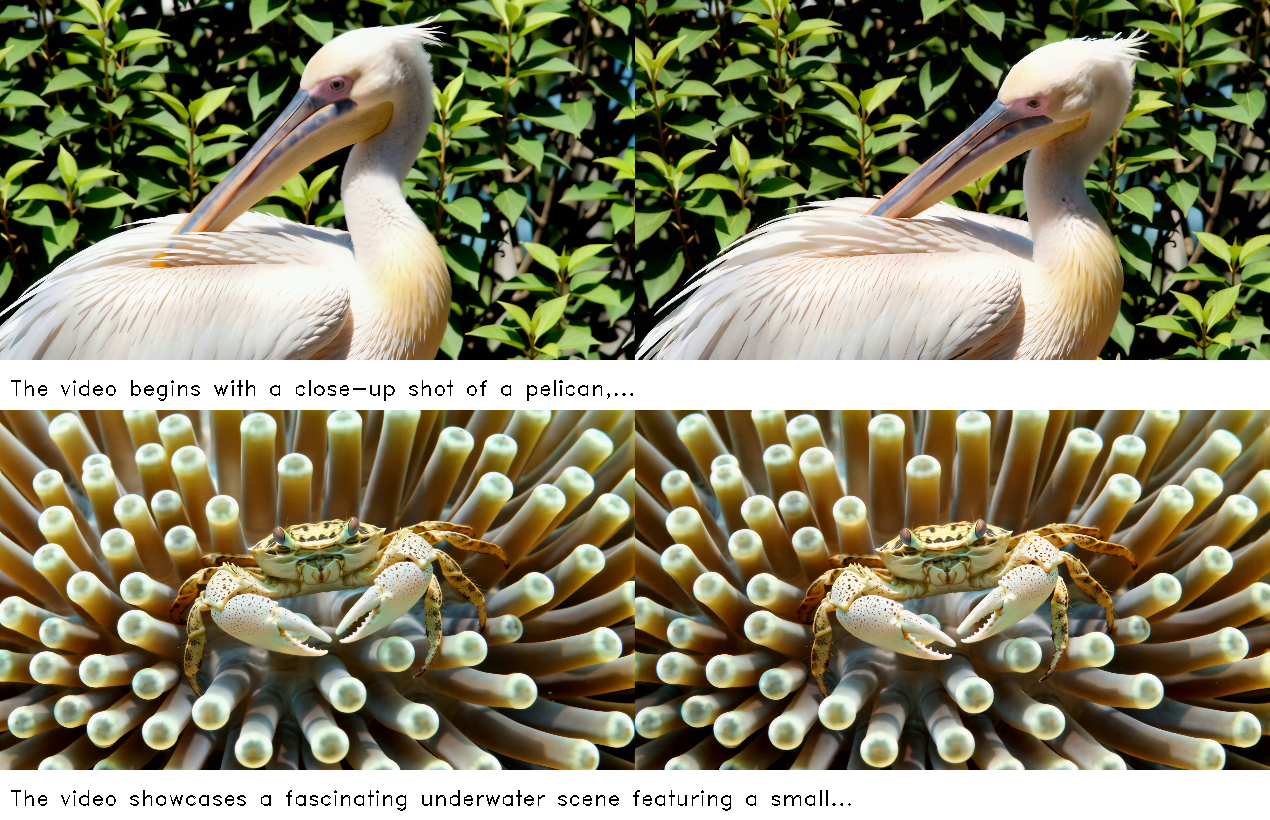}
    \caption{8K 29 Frames results: Each results spans for two rows. The frame indices are 0 and 28.}
    \label{fig:8k-more}
\end{figure}
\section{Extended Implementation Details}
\label{sec:supp_implementation_detail}

\subsection{Training Configuration}
We build~\ourwork on top of Wan2.1-T2V-1.3B~\cite{wan2025wan} and train both
stages with LoRA rank~16 using the AdamW optimizer (learning rate $1\!\times\!10^{-4}$,
$\beta_1{=}0.9$, $\beta_2{=}0.95$, weight decay $0.01$). All training is
conducted on $2\times$ NVIDIA RTX Pro 6000 (Ada) GPUs, batch size $1$ per GPU
with gradient accumulation $4$ (effective batch size $8$) for $15$K iterations.
Mixed-precision (\texttt{bf16}) is used throughout, and we adopt a flow-matching
objective consistent with the Wan2.1 base model.

\noindent\textbf{Stage~1 (semantic generator).} We finetune the base model with
temporal-scale RoPE ($r_t{=}4$) on $720\text{P}\times 81$-frame clips
sub-sampled at $4$~fps, so that an $81$-frame proxy spans an effective horizon
of $\sim\!20$ seconds at $16$~fps target frame rate.

\noindent\textbf{Stage~2 (detail branch).} The latent volume is partitioned
into spatiotemporal cubes of pixel size $(256,256,32)$, equivalent to
$(8,8,4)$ in the $4$D latent grid produced by the Wan2.1 3D-VAE. Border cubes
that cannot be evenly tiled adopt a smaller block size automatically. We
implement the asymmetric mask~$M$ (\S\ref{sec-joint-denoising}) on top of
PyTorch \texttt{flex\_attention} so that the kernel supports dynamic resolution
and length without re-compilation per shape.

\noindent\textbf{Optional 4K refinement.} The optional FFN refinement stage
finetunes only the feed-forward layers of the detail branch on $13$-frame
$3840\!\times\!2160$ clips drawn from UltraVideo~\cite{xue2025ultravideo}, for
$2$K iterations at the same learning rate.

\subsection{Inference Configuration}
Unless stated otherwise, all reported numbers and figures use Euler sampler and \emph{disable} classifier-free guidance (CFG), which is
justified by the ablation in (\S\ref{sec:ablation study} of the main paper. We perform inference on a
single RTX Pro 6000. End-to-end runtime includes text encoding, both denoising
stages, and 3D-VAE decoding.

\section{Detailed Metric Definitions}
\label{sec:supp_metrics}

This section formalises the metrics referenced in
Table~\ref{tab:quantitative_4k} and the ablation tables of the main paper.
For all metrics, $v\in\mathbb{R}^{T\times H\times W\times 3}$ denotes a video.
Higher-is-better metrics are marked $\uparrow$ and lower-is-better $\downarrow$.

\subsection{High-Definition Metrics}
The HD-FVD, HD-LPIPS, and HD-MSE metrics are designed for ultra-high-resolution
video evaluation, where standard FVD/LPIPS are constrained by the
$224\!\times\!224$ input size of their backbones. We follow the formulation
introduced by UltraGen~\cite{hu2026ultragen}.

\noindent\textbf{HD-FVD~$\downarrow$} Standard FVD computes the Fr\'echet
distance between I3D~\cite{carreira2017quo} feature distributions of generated
and reference videos. Because I3D operates on $224\!\times\!224$ inputs, naively
resizing a 4K clip to $224^2$ destroys all high-frequency content. HD-FVD
instead splits each frame into a regular grid of non-overlapping
$H_l\!\times\!W_l$ patches with $H_l\!\approx\! W_l\!\approx\! 224$, extracts
I3D features per spatiotemporal patch tube, and reports
\[
\mathrm{HD\text{-}FVD}
= \mathrm{FD}\!\left(\,\mathcal{N}(\mu_g,\Sigma_g)\,\Big\|\,\mathcal{N}(\mu_r,\Sigma_r)\right),
\]
where $(\mu_g,\Sigma_g)$ and $(\mu_r,\Sigma_r)$ are the means and covariances
of generated and reference patch-feature distributions, and
$\mathrm{FD}(\cdot\Vert\cdot)$ is the Fr\'echet distance. We use the public
Kinetics-pretrained I3D checkpoint and patch each $T\!=\!16$-frame clip into
$3\!\times\!3$ spatial patches at 4K (covering the full frame).

Specifically, we apply the same evaluation algorithm to both Wan2.1 (720p) and UltraGen (1080p), which results in relatively high HD-FVD scores. This is because the patch distributions of low-resolution and high-resolution videos differ significantly.

\noindent\textbf{HD-LPIPS~$\uparrow$} This \emph{no-reference} metric
quantifies how much high-frequency content is preserved in a generated video
by measuring its perceptual distance from progressively low-pass-filtered
copies of itself. Let $v_t$ denote the $t$-th frame of a video $V$ of length
$T$, and let $v_{t,D,2^k}$ denote $v_t$ bilinearly down-sampled by a factor
of $2^k$ and then bilinearly up-sampled back to the original resolution.
We define
\[
\mathrm{HD\text{-}LPIPS}(V) \;=\; \frac{1}{T}\sum_{t=1}^{T}\sum_{k\in\mathcal{K}}
\mathrm{LPIPS}\!\left(v_t,\, v_{t,D,2^k}\right),
\]
where $\mathcal{K}=\{3,4,5\}$ corresponds to down-sampling factors of $8$,
$16$, and $32$, chosen to capture detail across multiple spatial scales while
avoiding both noise-dominated (small factors) and structure-dominated (large
factors) regimes. A higher HD-LPIPS indicates that the generated video
differs more from its low-pass-filtered versions, suggesting richer
high-frequency content. We note that this metric reflects high-frequency
energy in general, and should therefore be interpreted alongside HD-FVD and
perceptual quality metrics to disambiguate genuine fine detail from
high-frequency artifacts.

\subsection{Text Video Alignment}
\noindent\textbf{CLIP~$\uparrow$.} We report frame-averaged cosine similarity
between the prompt embedding and per-frame visual embeddings. For a video with $T$ frames and
prompt $p$,
\[
\mathrm{CLIP}(v,p)\;=\;\tfrac{1}{T}\sum_{t=1}^{T}
\frac{\langle f_{\text{img}}(v_t),\, f_{\text{txt}}(p)\rangle}
{\|f_{\text{img}}(v_t)\|\,\|f_{\text{txt}}(p)\|}.
\]

\subsection{VBench Dimensions}
We follow the VBench~\cite{huang2023vbench} protocol and report the eight
dimensions used in Table~\ref{tab:quantitative_4k}. All scores lie in $[0,1]$
and higher is better.
\begin{itemize}
\setlength\itemsep{1pt}
\item \textbf{Subject Consistency (SC)} — frame-to-frame DINO~\cite{caron2021emerging}
cosine similarity of the foreground subject; measures whether the subject
identity is preserved across the clip.
\item \textbf{Background Consistency (BC)} — frame-to-frame CLIP image
cosine similarity of the background region; measures scene stability.
\item \textbf{Temporal Flickering (TF)} — mean absolute pixel difference
between consecutive frames in static regions; rewards low flicker.
\item \textbf{Motion Smoothness (MS)} — drops every other frame, interpolates
it back with the AMT~\cite{li2023amt} video frame interpolation model, and
reports the inverse reconstruction error; rewards physically plausible motion.
\item \textbf{Dynamic Degree (DD)} — average RAFT~\cite{teed2020raft} optical-flow
magnitude; rewards non-static videos. Reported in $[0,100]$ following the
official VBench convention.
\item \textbf{Aesthetic Quality (AQ)} — per-frame LAION aesthetic predictor
score, averaged temporally.
\item \textbf{Imaging Quality (IQ)} — per-frame MUSIQ~\cite{ke2021musiq}
SPAQ-trained image-quality predictor, averaged temporally.
\item \textbf{Color (Clr.)} — GRiT~\cite{wu2024grit}
caption-based color attribute alignment between the rendered subject and the
color word in the prompt.
\end{itemize}

\section{Additional Ablation Study on CFG Removal}
\begin{table}[h]
\centering
\caption{Ablation on the removal of classifier-free guidance (CFG).}
\label{tab:ablation_cfg}
\setlength{\tabcolsep}{8pt}
\renewcommand{\arraystretch}{1.15}
\begin{tabular}{lcccc}
\toprule
\textbf{Variant}
& \textbf{Inference}
& \textbf{FVD}$\downarrow$
& \textbf{HD-LPIPS}$\uparrow$
& \textbf{CLIP}$\uparrow$ \\
\midrule
w/ 4K fine-tuning  & w/ CFG  & 273.64 & 0.6612 & 0.3012 \\
w/ 4K fine-tuning  & w/o CFG & 284.71 & 0.6695 & 0.3019 \\
w/o 4K fine-tuning & w/ CFG  & 408.53 & 0.6398 & 0.3017 \\
w/o 4K fine-tuning & w/o CFG & 409.15 & 0.6401 & 0.3011 \\
\bottomrule
\end{tabular}
\end{table}

\noindent\textbf{Ablation on removing classifier-free guidance (CFG).}
Since the global proxy already provides a deterministic semantic target for high-resolution denoising, we remove classifier-free guidance to improve inference efficiency. We evaluate this design both with and without optional 4K fine-tuning. As shown in Table~\ref{tab:ablation_cfg}, disabling CFG has no significant impact on the quantitative metrics in either setting. Meanwhile, it nearly halves the DiT inference cost by eliminating the unconditional generation branch. We therefore disable CFG in our final generation pipeline.
\section{Broader Impact}
\label{sec:supp_broader_impact}

\noindent\textbf{Positive societal impacts.}
\ourwork lowers the barrier to ultra-high-resolution long-video generation by
training on as few as two consumer-class GPUs (RTX Pro 6000) at 720P resolution
and transferring directly to 4K and beyond. This has several positive
implications. (i) It democratises high-resolution video synthesis, making it
practical for academic groups, independent creators, and non-profit
educational projects that cannot afford clusters of $32$--$64$ H-class GPUs.
(ii) It reduces the energy and carbon footprint of training and serving UHR
video models by an order of magnitude, since the dominant cost --- quadratic
spatiotemporal attention --- is replaced with a locality-preserving variant
whose attention FLOPs grow linearly in the number of spatiotemporal cubes.
(iii) It can support accessibility applications such as automatic generation
of high-quality educational visuals, sign-language interpretation videos, or
visualisations for the visually-impaired community where high spatial detail
matters.

\noindent\textbf{Potential negative societal impacts.}
Like all powerful generative video systems,~\ourwork could in principle be
misused for (i) producing photorealistic disinformation or non-consensual
synthetic media (``deepfakes''), (ii) generating misleading long-form footage
of public figures or events, or (iii) bypassing moderation tools that were
trained on lower-resolution synthetic content. Because our framework is
specifically designed to scale long-horizon and ultra-high-resolution
synthesis, the resulting outputs are harder to distinguish from real footage
than those of short low-resolution generators, which sharpens the dual-use
concern.

\newpage

\end{document}